\documentclass{article}

\usepackage{arxiv}

\usepackage[utf8]{inputenc} 
\usepackage[T1]{fontenc}    
\usepackage{hyperref}       
\usepackage{url}            
\usepackage{booktabs}       
\usepackage{amsfonts}       
\usepackage{nicefrac}       
\usepackage{microtype}      
\usepackage{lipsum}		
\usepackage{graphicx}
\usepackage{natbib}
\usepackage{doi}

\usepackage{amsmath}
\usepackage{amssymb}
\usepackage{caption}
\usepackage{subcaption}
\usepackage{multirow}
\usepackage{longtable}
\usepackage{tabu}
\usepackage{pdflscape} 
\usepackage{xltabular}

\title{On the robustness of self-supervised representations for multi-view object classification}

\author{ \href{https://orcid.org/0000-0003-2822-7146}{\includegraphics[scale=0.06]{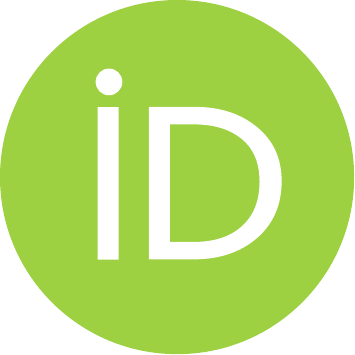}\hspace{1mm}David ~Torpey}\\
	School of Computer Science and Applied Mathematics\\
	University of the Witwatersrand\\
	Johannesburg, South Africa \\
	\texttt{674425@students.wits.ac.za} \\
	\And
	\href{https://orcid.org/0000-0003-0783-2072}{\includegraphics[scale=0.06]{orcid.pdf}\hspace{1mm}Richard ~Klein} \\
	School of Computer Science and Applied Mathematics\\
	University of the Witwatersrand\\
	Johannesburg, South Africa \\
	\texttt{richard.klein@wits.ac.za} \\
}

\renewcommand{\shorttitle}{Self-supervised representation multi-view robustness}

\hypersetup{
pdfauthor={David ~Torpey, Richard ~Klein},
}

\begin{document}
\maketitle

\begin{abstract}
	It is known that representations from self-supervised pre-training can perform on par, and often better, on various downstream tasks than representations from fully-supervised pre-training. This has been shown in a host of settings such as generic object classification and detection, semantic segmentation, and image retrieval. However, some issues have recently come to the fore that demonstrate some of the failure modes of self-supervised representations, such as performance on non-ImageNet-like data, or complex scenes. In this paper, we show that self-supervised representations based on the instance discrimination objective lead to better representations of objects that are more robust to changes in the viewpoint and perspective of the object. We perform experiments of modern self-supervised methods against multiple supervised baselines to demonstrate this, including approximating object viewpoint variation through homographies, and real-world tests based on several multi-view datasets. We find that self-supervised representations are more robust to object viewpoint and appear to encode more pertinent information about objects that facilitate the recognition of objects from novel views.
\end{abstract}


In recent times, self-supervised learning (SSL) has come to the fore as a viable alternative to fully-supervised models as a way to perform large-scale pre-training that can transfer successfully to downstream tasks \citep{simclr,simclrv2,dino,swav}. It is known that self-supervised (SS) pre-training results in performance that is on par, and often superior, to supervised pre-training \citep{Ericsson2021HowTransfer} (usually by pre-training on the ImageNet \citep{imagenet} dataset). However, some of the failure modes are 1) poor visual grounding resulting in decreased performance on scene images versus object-centric images \citep{selvaraju2020casting}, and 2) less robustness than supervised models when transferring to data with a distribution shift from ImageNet \citep{Ericsson2021HowTransfer}, such as medical imaging or satellite imagery. However, very little is known about self-supervised representations, their pros and cons, what they do and do not encode, and potential ways to improve them. It is of crucial importance to understand these models and when/where they perform well so that we can lay a path forward to improve on them.

On a high level, we attempt to delve deeper into the efficacy of SS representations, and what they potentially encode differently from supervised representations - specifically in terms of robustness and invariance to viewpoint. Viewpoint invariance is a well-studied area of research, and there is biological evidence that suggests that it is a core trait of the human visual system \citep{viewpointpsyc}. Firstly, we perform an analysis of SSL versus supervised learning in the context of controlled changes in object view by applying homographies of varying strengths to the images. We perform this analysis on datasets with different properties (e.g. fine-grained/generic, high/low cardinality label set). These homographies approximate possible object transformations in the real world and thus serve as a good proxy to test this hypothesis on large scale, diverse datasets that are not inherently multi-view.

We then perform a set of experiments on real-world, multi-view scenarios. These datasets cover a variety of object types, scenes, and scenarios that could be seen in the wild. The primary goal of this analysis is to empirically demonstrate that SS representations encode information that provides for a better ability to generalise to novel object views in realistic settings (i.e. datasets that are inherently multi-view). The secondary goal of this analysis is to verify the finding of the previous experiment in which the different object views are synthetically generated through homographies. This analysis is achieved through nearest neighbour searches directly in the respective networks' feature spaces (i.e. there is no form of weight adaptation to the backbone network). We perform ablations to show that the results hold even with these adaptations. Our contributions can be summarised as follows.

    (1) Performing an analysis comparing SS representations to supervised representations in terms of robustness to viewpoint changes using common image classification benchmark datasets from the literature. This is done by approximating changes to object view with homographies, and it enables the quantifying of performance in a \emph{controlled environment}.
    (2) Performing a similar analysis to the above using multiple real-world, multi-view datasets to quantify performance in the wild. These are the more important experiments as they demonstrate realistic multi-view scenarios.
    (3) Demonstrate that SS representations are more robust to both A) approximate/synthetic viewpoint variations (homography) and B) real-world viewpoint variations through multiple multi-view datasets.
    (4) Demonstrate that SS representations require less context than multiple supervised baselines to identify an object using the embeddings directly from the backbone networks.


The rest of the paper is formatted as follows. Section \ref{sec:related_work} presents related work in the domain, including modern SS techniques and previous work analysing SS representations. Section \ref{sec:methodology} presents our research methodology. Section \ref{sec:experiments} presents an experimental analysis, including experiments on large-scale datasets using homographies to approximate viewpoint changes as well as tests on various real-world multi-view datasets. Finally, Section \ref{sec:conclusion} concludes the research with some final remarks.

\section{Related Work}
\label{sec:related_work}
A wealth of previous work exists in the realm of SSL \citep{swav,moco,simclr,byol,color,contextpred}. We discuss the methods pertinent to modern SSL, and where our work fits into the domain. It should be noted that approaches to SSL can largely be cast as either generative (e.g. generative adversarial networks \citep{gan} and variational autoencoders \citet{vae}) or discriminative (pretext tasks and instance discrimination (ID)/contrastive learning (CL)). This review of prior work focuses on discriminative approaches, as these are the most popular techniques currently.

\subsection{Pretext Tasks}
SSL can be characterised as a way of obtaining a supervision signal from the input data itself. One way to achieve this is by defining a so-called \emph{pretext task}, which is a proxy task for which labels are derived from the input images. A huge variety of such pretext tasks have been proposed including relative patch prediction \citep{contextpred}, predicting rotations \citep{rotfeatdecoup,rotpred}, jigsaw puzzles \citep{jigsaw}, colourisation \citep{colourisation1,colourisation2,colourisation3}, inpainting \citep{inpainting}, and others \citep{splitbrain,learningtocount}.

For example, \citet{contextpred} define a proxy task of predicting the position on a neighbouring image patch relative to the central patch on a $3 \times 3$ grid of patches. \citet{jigsaw} train a network to solve a jigsaw puzzle by separating the image into numbered patches, shuffling these patches, and training the network to reassemble the patches into the correct order. One of the more successful pretext tasks is colourisation \citep{colourisation1,colourisation2}, in which a network is trained to predict the colourised version of an input grayscale image.

Many early approaches to SSL were based on such hand-crafted pretext tasks. However, defining a pretext task that can learn generic, transferable image representations for downstream tasks is difficult. More recently, approaches based on \emph{instance discrimination} have seen more use and gained more popularity from the research community, for various reasons including improved scalability, and better-learned representations that transfer better to various downstream tasks.

\subsection{Instance Discrimination}
The core idea of ID is to make each image its own class, thereby tasking the network with discriminating between individual images (i.e. instances), instead of the more common discrimination between classes. In its naive formulation, this would introduce a natural bottleneck when the dataset grows too large, as the number of classes would grow linearly with it. Thus, it would not be possible to use the typical parametric cross-entropy loss. To make this problem tractable, \citet{nonparametric} introduce a non-parametric softmax classifier that is trained to be instance-discriminative. Noise-contrastive estimation (NCE) \citep{nce} is used to make computing the non-parametric softmax more efficient. \citet{parametricid} manage to perform parametric ID for SSL.

\citet{cpc} propose a similar loss function based on NCE known as InfoNCE. This loss has been widely used in the literature \citep{simclr,simclrv2}. Their model is based on the idea of predictive coding \citep{predictivecoding}, and predicts the future in latent space using an autoregressive model.

Many SSL methods fall into a class known as \emph{contrastive learning} \citep{simclr,simclrv2,cpc,moco,mocov2}. \citet{simclr} propose a general framework for SSL based on CL. Two networks, in a Siamese-like fashion, operate on two distinct views of the same input image. These views are generated by random data augmentation, such as cropping, colour dropping, colour jitter, and blurring. This random view generation through augmentation is a common approach in the literature. The InfoNCE between the embeddings of the two views is then minimised. However, such an architecture requires a very large batch size (e.g. 4096+) to be effective, since sufficiently many negative samples need to be present for the CL to work.

Various architectures that overcome the need for such large batch sizes have been proposed \citep{byol,barlowtwins,swav}. \citet{byol} propose a student-teacher paradigm in which the student network predicts the embedding of the teacher network. Only the weights of the student are updated via backpropagation, whereas teacher weights are updated using an exponential moving average.

Many other ID approaches have been proposed, including specialised architectures such as memory banks \citep{moco,mocov2}, clustering-based approaches \citep{swav,deepcluster}, mutual information maximisation \citep{mutualinfomax}, vision transformer-based models \citep{dino}, and combinations of pretext tasks and CL \citep{pirl}. However, many of those have bottlenecks inhibiting their scalability and widespread use, including complicated architectures, training regimes, and very large compute requirements.

\subsection{Issues with Self-Supervised Representations}
It has been shown that for many downstream tasks (i.e. image classification, object detection, semantic segmentation, surface normal estimation, and image retrieval), SS representations consistently outperform supervised representations \citep{Ericsson2021HowTransfer}. However, there are some issues with SSL and the associated learned representations. Firstly, it has been shown that their efficacy when the downstream task has a large distribution shift from ImageNet is much lower, and their comparative performance versus a similarly pre-trained supervised alternative is poor.

Secondly, it has been shown SS models have poor visual grounding \citep{selvaraju2020casting}, resulting in worse performance on images that are not object-centric, such as scene images with many objects. A GradCAM-based \citep{gradcam} loss is proposed to overcome this poor visual grounding.

Interestingly, \citet{demyst} find that SS models fail to capture viewpoint and category instance invariance. However, this does not affect our findings of SSL models being more robust to viewpoint, since there are a few key differences from our work. Firstly, we quantify performance in a different way (we measure based on the common linear evaluation paradigm, whereas they use an explicit measure of invariance based on the proposal in \citep{measureinvariance}). Secondly, we test on more varied and realistic datasets specifically tailored to viewpoint. Lastly, we study a different class of models (i.e. SWaV \citep{swav} and DINO \citep{dino} instead of MoCo \citep{moco} and PIRL \citep{pirl}). These are completely different architectures trained in very different ways, resulting in inherently different properties. Furthermore, the findings for the slightly worse performance of SSL versus supervised in terms of viewpoint invariance can partially be attributed to the nature of the datasets used (e.g. ALOI \citep{aloi}), which contains images that are very dissimilar to ImageNet, which has previously been shown to result in poor performance in SSL \citep{Ericsson2021HowTransfer}. In fact, we too observe similar findings for datasets with large distribution shift from ImageNet.


\section{Methodology}
\label{sec:methodology}
\subsection{Measuring Robustness to Viewpoint Variation}
Consider functions $f : \mathcal{X} \rightarrow \mathbb{R}^n$ and $g : \mathcal{X} \rightarrow \mathbb{R}^n$, and a sample space of images $\mathcal{X}$. These are a supervised pre-trained model, and a self-supervised pre-trained model, respectively.

We aim to analyse the efficacy and representational power of embeddings $f(x), g(x) \in \mathbb{R}^n$ in terms of robustness to viewpoint variation on a host of different datasets. We assume that both $f$ and $g$ have been estimated by pre-training on ImageNet (as this is the common paradigm in the literature). We then employ a transfer learning approach to analyse the pre-trained models and resultant features.

Essentially, we aim to demonstrate empirically that the SS representations produced by $g$ are more robust to viewpoint changes than those from $f$. Mathematically, this can be formalised as follows. Consider a function $V : \mathcal{X} \rightarrow \mathcal{X}$ that is tasked with changing an object's viewpoint. This could be a homography, affine transformation, or in general any transformation that can perturb an object along one or more of its axes of variation (such as those that would occur in the wild). Then, a function $g$ is more robust to $V$ than a function $f$, if:
\begin{equation} \label{eqn:criterion}
    \mathbb{E}[L(f(x), f(V(x)))] \ge \mathbb{E}[L(g(x), g(V(x)))]
\end{equation}
for some loss function $L$, and for all $x \in \mathcal{X}$. \emph{This is the criterion we use to measure and compare models' multi-view recognition performance, and implicitly, their viewpoint invariance}. Ideally, we would want full invariance to $V$, but this is not realistic in practice. However, there are scenarios where full invariance is not ideal for other classes of perturbations besides viewpoint. For example, invariance to colour will make a downstream task of colour classification difficult.

\subsection{Synthetic Viewpoint Variation Analysis}
\begin{figure}
    \centering
    \includegraphics[width=0.99\linewidth]{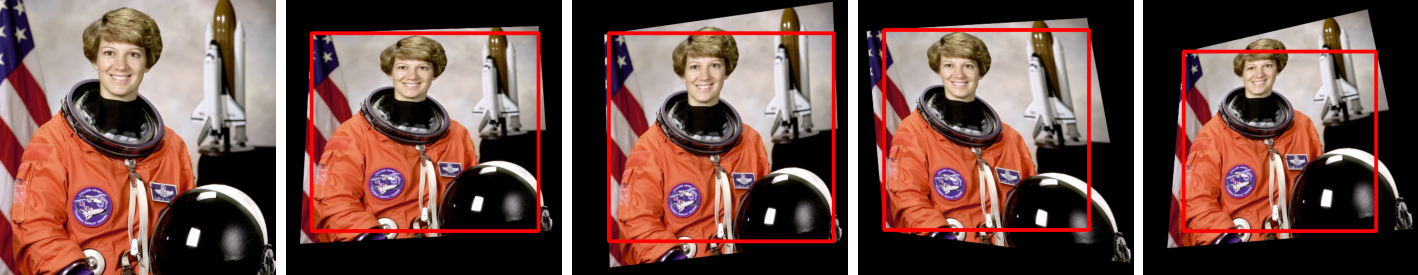}
    \caption{Homographies of varying strengths (from left: $H_{0.0}$, $H_{0.2}$, $H_{0.4}$, $H_{0.6}$, and $H_{0.8}$). The red boxes depict our \emph{bounded homography}.}
    \label{fig:homog_viz}
\end{figure}

To analyse the robustness of these supervised and SS representations to viewpoint variation, we opt for a two-stage approach. We first investigate the viewpoint invariance in a controlled environment, by synthetically varying the view of an object by applying a random homography \emph{to the testing images}. We apply two kinds of homographies: a regular homography where any new background created is left in the image (the default for many libraries), and a \emph{bounded homography}, whereby we crop the maximum-area inscribed axis-aligned rectangle from the resulting polygon. This bounded homography enables the ability to test whether the black background in a default homography affects performance and if the models are biased toward this black background (see Fig. \ref{fig:homog_viz} for a comparison).

We represent a homography as $H_{\alpha} : \mathcal{X} \rightarrow \mathcal{X}$, where $\alpha \in [0, 1]$ is a factor controlling the strength of the homography. The goal of this set of experiments is to have full control of the amount of viewpoint variation through the strength of the homography, which allows for a systematic analysis of the two modelling paradigms.

For the linear evaluation experiments, we train a multinomial logistic regression model on the representations $f(x)$ and $g(x)$ (using the L-BFGS optimiser) on a training dataset $X_{\text{tr}} \subset \mathcal{X}$ with labels $Y$. We then compute accuracy on the test set with a homography applied to each image: $\{H_{\alpha}(x) | x \in X_{\text{te}}\}$. The application of this homography serves as a mechanism to obtain novel object views, as homographies can be used to approximate object transformations seen in the real world.

\subsection{Real-World Viewpoint Variation Analysis}
Motivated by the results on synthetic data, we also perform experiments on real-world, multi-view datasets using the representations directly from $f$ and $g$ (i.e. without training a classifier on top of $f$ or $g$). Importantly, these experiments serve as the main contribution of the paper, as they measure the efficacy of the 2 modelling paradigms in a realistic scenario through the use of real-world datasets. The viewpoint invariance is measured in the form of a $k$-nearest neighbour search, with $k = 1$. This paradigm allows for better empirical evidence as to whether the bare representations (without a classifier learning from them) are encoding viewpoint in the way we hypothesise. For these real-world experiments, we use a suite of datasets that were explicitly curated to test multi-view performance, and that cover a wide range of object and task types.

\section{Experiments}
\label{sec:experiments}
\subsection{Experimental Setup}
\subsubsection{Datasets}
We use the following datasets for the deformation experiments in which the homography is applied to the testing images: Aircraft \citep{aircraft}, Birdsnap \citep{birdsnap}, Color \citep{color}, Caltech101 \citep{caltech101}, CIFAR10 \citep{cifar10}, CIFAR100 \citep{cifar100}, Food \citep{food}. These datasets are used to be consistent with previous work on benchmarking SS representations for downstream image classification.

For the real-world, multi-view experiments, we use multiple realistic datasets, including the multi-view car (MVC) dataset \citep{mvc}, a multi-view stereo dataset of generic objects (Recon3D) \citep{3drecon}, a multi-view, multi-class dataset \citep{mvmc} (MVMC), the Amsterdam library of object images \citep{aloi} (ALOI), and a multi-view stereo face dataset \citep{stereoface} (Stereo Face). The latter three of these datasets are from the EPFL CVLAB data repository\footnote{https://www.epfl.ch/labs/cvlab/data/}. These datasets present a diverse range of multi-view scenarios, object and background types, and difficulty for the models under assessment.

\subsubsection{Models}
In order to effectively evaluate the efficacy of SS and supervised representations for invariance to viewpoint, we test a suite of a total of 15 different models. We separate these 15 models into 3 groups: 1) SSL (ID) - ID-based SSL, 2) SSL (PT) - pretext task-based SSL, and 3) supervised. The models within each group are: \textbf{SSL (ID)}: SWaV (RN50) and SWaV (RN50w2) \citep{swav}, SimCLR (RN50) and SimCLR (RN50w2) \citep{simclr,simclrv2}, DINO (ViT) and DINO (RN50) \citep{dino}, MoCo v2 (RN50) \citep{mocov2}, Barlow Twins (RN50) \citep{barlowtwins}; \textbf{SSL (PT)}: RotNet (RN50) \citep{rotpred}, Jigsaw (RN50) \citep{jigsaw}, Colorization (RN50) \citep{colourisation1}; \textbf{Supervised}: RN50 \citep{resnet}, ViT \citep{vit}, EfficientNet \citep{effnet}, RegNet \citep{regnet}. This suite of models provides a wide variety of SSL methods, backbone architectures, and SSL paradigms (ID, CL, pre-text tasks). All models are pre-trained on ImageNet, and available in PyTorch Hub\footnote{https://pytorch.org/hub/} and the VISSL Model Zoo\footnote{https://github.com/facebookresearch/vissl/blob/main/MODEL\_ZOO.md}.


\subsection{Robustness to Deformation} \label{sec:synthetic}

We first conduct experiments on common large-scale downstream benchmark datasets to quantify the performance of the models in the multi-view case by approximating different object views and perspectives by applying a homography to each testing image. We use the linear evaluation paradigm \citep{simclr,byol,moco,simclrv2} from the literature to compute our metrics for this set of experiments.

\begin{table*}[!h]
\caption{\label{tbl:homog_results}Linear evaluation performance on common image classification benchmarks for increasing homography strengths (top 3 best-performing models). Results are averaged over 10 random trials.}
\centering
\resizebox{\columnwidth}{!}{%
\begin{tabular}{l|lll||lll||lll||lll||lll|}
\cline{2-16}
                                                  & \multicolumn{3}{c|}{$H_{0.0}$}                                                          & \multicolumn{3}{c|}{$H_{0.2}$}                                                            & \multicolumn{3}{c|}{$H_{0.4}$}                                                            & \multicolumn{3}{c|}{$H_{0.6}$}                                                            & \multicolumn{3}{c|}{$H_{0.8}$}                                                            \\ \hline
\multicolumn{1}{|l|}{\multirow{3}{*}{CIFAR10}}    & \multicolumn{1}{l|}{Supervised (ViT)}  & \multicolumn{1}{l|}{Supervised} & 92.29 & \multicolumn{1}{l|}{Supervised (ViT)}    & \multicolumn{1}{l|}{Supervised} & 91.68 & \multicolumn{1}{l|}{Supervised (ViT)}    & \multicolumn{1}{l|}{Supervised} & 90.94 & \multicolumn{1}{l|}{Supervised (ViT)}    & \multicolumn{1}{l|}{Supervised} & 88.7  & \multicolumn{1}{l|}{Supervised (ViT)}    & \multicolumn{1}{l|}{Supervised} & 79.94 \\ \cline{2-16} 
\multicolumn{1}{|l|}{}                            & \multicolumn{1}{l|}{SWaV (RN50w2)}     & \multicolumn{1}{l|}{SSL (ID)}   & 91.52 & \multicolumn{1}{l|}{DINO (ViT)}          & \multicolumn{1}{l|}{SSL (ID)}   & 89.04 & \multicolumn{1}{l|}{DINO (ViT)}          & \multicolumn{1}{l|}{SSL (ID)}   & 83.67 & \multicolumn{1}{l|}{Supervised EffNet}   & \multicolumn{1}{l|}{Supervised} & 74.87 & \multicolumn{1}{l|}{Supervised EffNet}   & \multicolumn{1}{l|}{Supervised} & 67.83 \\ \cline{2-16} 
\multicolumn{1}{|l|}{}                            & \multicolumn{1}{l|}{DINO (ViT)}        & \multicolumn{1}{l|}{SSL (ID)}   & 91.12 & \multicolumn{1}{l|}{Barlow Twins (RN50)} & \multicolumn{1}{l|}{SSL (ID)}   & 84.01 & \multicolumn{1}{l|}{Barlow Twins (RN50)} & \multicolumn{1}{l|}{SSL (ID)}   & 79.92 & \multicolumn{1}{l|}{DINO (ViT)}          & \multicolumn{1}{l|}{SSL (ID)}   & 74.15 & \multicolumn{1}{l|}{DINO (ViT)}          & \multicolumn{1}{l|}{SSL (ID)}   & 64.65 \\ \hline \hline
\multicolumn{1}{|l|}{\multirow{3}{*}{CIFAR100}}   & \multicolumn{1}{l|}{Supervised (ViT)}  & \multicolumn{1}{l|}{Supervised} & 75.43 & \multicolumn{1}{l|}{Supervised (ViT)}    & \multicolumn{1}{l|}{Supervised} & 73.99 & \multicolumn{1}{l|}{Supervised (ViT)}    & \multicolumn{1}{l|}{Supervised} & 72.67 & \multicolumn{1}{l|}{Supervised (ViT)}    & \multicolumn{1}{l|}{Supervised} & 68.71 & \multicolumn{1}{l|}{Supervised (ViT)}    & \multicolumn{1}{l|}{Supervised} & 59.73 \\ \cline{2-16} 
\multicolumn{1}{|l|}{}                            & \multicolumn{1}{l|}{SWaV (RN50w2)}     & \multicolumn{1}{l|}{SSL (ID)}   & 74.65 & \multicolumn{1}{l|}{DINO (ViT)}          & \multicolumn{1}{l|}{SSL (ID)}   & 68.66 & \multicolumn{1}{l|}{DINO (ViT)}          & \multicolumn{1}{l|}{SSL (ID)}   & 62.82 & \multicolumn{1}{l|}{DINO (ViT)}          & \multicolumn{1}{l|}{SSL (ID)}   & 55.26 & \multicolumn{1}{l|}{Supervised EffNet}   & \multicolumn{1}{l|}{Supervised} & 48.22 \\ \cline{2-16} 
\multicolumn{1}{|l|}{}                            & \multicolumn{1}{l|}{DINO (ViT)}        & \multicolumn{1}{l|}{SSL (ID)}   & 74.43 & \multicolumn{1}{l|}{Barlow Twins (RN50)} & \multicolumn{1}{l|}{SSL (ID)}   & 63.03 & \multicolumn{1}{l|}{Supervised EffNet}   & \multicolumn{1}{l|}{Supervised} & 58.25 & \multicolumn{1}{l|}{Supervised EffNet}   & \multicolumn{1}{l|}{Supervised} & 54.36 & \multicolumn{1}{l|}{DINO (ViT)}          & \multicolumn{1}{l|}{SSL (ID)}   & 48.2  \\ \hline \hline
\multicolumn{1}{|l|}{\multirow{3}{*}{Color}}      & \multicolumn{1}{l|}{SWaV (RN50w2)}     & \multicolumn{1}{l|}{SSL (ID)}   & 94.04 & \multicolumn{1}{l|}{DINO (ViT)}          & \multicolumn{1}{l|}{SSL (ID)}   & 91.94 & \multicolumn{1}{l|}{Supervised (ViT)}    & \multicolumn{1}{l|}{Supervised} & 90.75 & \multicolumn{1}{l|}{Supervised (ViT)}    & \multicolumn{1}{l|}{Supervised} & 88.64 & \multicolumn{1}{l|}{Supervised (ViT)}    & \multicolumn{1}{l|}{Supervised} & 81.9  \\ \cline{2-16} 
\multicolumn{1}{|l|}{}                            & \multicolumn{1}{l|}{DINO (ViT)}        & \multicolumn{1}{l|}{SSL (ID)}   & 93.78 & \multicolumn{1}{l|}{SWaV (RN50w2)}       & \multicolumn{1}{l|}{SSL (ID)}   & 91.66 & \multicolumn{1}{l|}{DINO (ViT)}          & \multicolumn{1}{l|}{SSL (ID)}   & 89.79 & \multicolumn{1}{l|}{DINO (ViT)}          & \multicolumn{1}{l|}{SSL (ID)}   & 86.03 & \multicolumn{1}{l|}{DINO (ViT)}          & \multicolumn{1}{l|}{SSL (ID)}   & 80.46 \\ \cline{2-16} 
\multicolumn{1}{|l|}{}                            & \multicolumn{1}{l|}{DINO (RN50)}       & \multicolumn{1}{l|}{SSL (ID)}   & 93.55 & \multicolumn{1}{l|}{Supervised (ViT)}    & \multicolumn{1}{l|}{Supervised} & 91.66 & \multicolumn{1}{l|}{SWaV (RN50w2)}       & \multicolumn{1}{l|}{SSL (ID)}   & 88.47 & \multicolumn{1}{l|}{SWaV (RN50w2)}       & \multicolumn{1}{l|}{SSL (ID)}   & 84.09 & \multicolumn{1}{l|}{SWaV (RN50w2)}       & \multicolumn{1}{l|}{SSL (ID)}   & 77.28 \\ \hline \hline
\multicolumn{1}{|l|}{\multirow{3}{*}{Caltech101}} & \multicolumn{1}{l|}{DINO (ViT)}        & \multicolumn{1}{l|}{SSL (ID)}   & 96.45 & \multicolumn{1}{l|}{Supervised (ViT)}    & \multicolumn{1}{l|}{Supervised} & 94.05 & \multicolumn{1}{l|}{DINO (ViT)}          & \multicolumn{1}{l|}{SSL (ID)}   & 93.31 & \multicolumn{1}{l|}{DINO (ViT)}          & \multicolumn{1}{l|}{SSL (ID)}   & 91.01 & \multicolumn{1}{l|}{Supervised (ViT)}    & \multicolumn{1}{l|}{Supervised} & 83.21 \\ \cline{2-16} 
\multicolumn{1}{|l|}{}                            & \multicolumn{1}{l|}{SWaV (RN50w2)}     & \multicolumn{1}{l|}{SSL (ID)}   & 95.34 & \multicolumn{1}{l|}{DINO (ViT)}          & \multicolumn{1}{l|}{SSL (ID)}   & 93.87 & \multicolumn{1}{l|}{Supervised (ViT)}    & \multicolumn{1}{l|}{Supervised} & 92.22 & \multicolumn{1}{l|}{Supervised (ViT)}    & \multicolumn{1}{l|}{Supervised} & 90.32 & \multicolumn{1}{l|}{DINO (ViT)}          & \multicolumn{1}{l|}{SSL (ID)}   & 81.66 \\ \cline{2-16} 
\multicolumn{1}{|l|}{}                            & \multicolumn{1}{l|}{DINO (RN50)}       & \multicolumn{1}{l|}{SSL (ID)}   & 95.02 & \multicolumn{1}{l|}{Supervised EffNet}   & \multicolumn{1}{l|}{Supervised} & 93.4  & \multicolumn{1}{l|}{Supervised EffNet}   & \multicolumn{1}{l|}{Supervised} & 91.86 & \multicolumn{1}{l|}{Supervised EffNet}   & \multicolumn{1}{l|}{Supervised} & 88.99 & \multicolumn{1}{l|}{Supervised EffNet}   & \multicolumn{1}{l|}{Supervised} & 81.52 \\ \hline \hline
\multicolumn{1}{|l|}{\multirow{3}{*}{Aircraft}}   & \multicolumn{1}{l|}{DINO (ViT)}        & \multicolumn{1}{l|}{SSL (ID)}   & 59.65 & \multicolumn{1}{l|}{DINO (ViT)}          & \multicolumn{1}{l|}{SSL (ID)}   & 53.95 & \multicolumn{1}{l|}{DINO (ViT)}          & \multicolumn{1}{l|}{SSL (ID)}   & 48.06 & \multicolumn{1}{l|}{DINO (ViT)}          & \multicolumn{1}{l|}{SSL (ID)}   & 41.95 & \multicolumn{1}{l|}{DINO (ViT)}          & \multicolumn{1}{l|}{SSL (ID)}   & 36.12 \\ \cline{2-16} 
\multicolumn{1}{|l|}{}                            & \multicolumn{1}{l|}{DINO (RN50)}       & \multicolumn{1}{l|}{SSL (ID)}   & 59.29 & \multicolumn{1}{l|}{DINO (RN50)}         & \multicolumn{1}{l|}{SSL (ID)}   & 47.72 & \multicolumn{1}{l|}{Barlow Twins (RN50)} & \multicolumn{1}{l|}{SSL (ID)}   & 41.62 & \multicolumn{1}{l|}{Barlow Twins (RN50)} & \multicolumn{1}{l|}{SSL (ID)}   & 35.8  & \multicolumn{1}{l|}{DINO (RN50)}         & \multicolumn{1}{l|}{SSL (ID)}   & 32.33 \\ \cline{2-16} 
\multicolumn{1}{|l|}{}                            & \multicolumn{1}{l|}{SWaV (RN50w2)}     & \multicolumn{1}{l|}{SSL (ID)}   & 56.14 & \multicolumn{1}{l|}{Barlow Twins (RN50)} & \multicolumn{1}{l|}{SSL (ID)}   & 47.58 & \multicolumn{1}{l|}{DINO (RN50)}         & \multicolumn{1}{l|}{SSL (ID)}   & 40.94 & \multicolumn{1}{l|}{DINO (RN50)}         & \multicolumn{1}{l|}{SSL (ID)}   & 35.57 & \multicolumn{1}{l|}{Barlow Twins (RN50)} & \multicolumn{1}{l|}{SSL (ID)}   & 31.76 \\ \hline \hline
\multicolumn{1}{|l|}{\multirow{3}{*}{Birdsnap}}   & \multicolumn{1}{l|}{DINO (ViT)}        & \multicolumn{1}{l|}{SSL (ID)}   & 59.55 & \multicolumn{1}{l|}{DINO (ViT)}          & \multicolumn{1}{l|}{SSL (ID)}   & 56.37 & \multicolumn{1}{l|}{DINO (ViT)}          & \multicolumn{1}{l|}{SSL (ID)}   & 53.54 & \multicolumn{1}{l|}{DINO (ViT)}          & \multicolumn{1}{l|}{SSL (ID)}   & 48.76 & \multicolumn{1}{l|}{DINO (ViT)}          & \multicolumn{1}{l|}{SSL (ID)}   & 41.18 \\ \cline{2-16} 
\multicolumn{1}{|l|}{}                            & \multicolumn{1}{l|}{Supervised (ViT)}  & \multicolumn{1}{l|}{Supervised} & 49.39 & \multicolumn{1}{l|}{Supervised (ViT)}    & \multicolumn{1}{l|}{Supervised} & 47.83 & \multicolumn{1}{l|}{Supervised (ViT)}    & \multicolumn{1}{l|}{Supervised} & 45.53 & \multicolumn{1}{l|}{Supervised (ViT)}    & \multicolumn{1}{l|}{Supervised} & 42.56 & \multicolumn{1}{l|}{Supervised (ViT)}    & \multicolumn{1}{l|}{Supervised} & 36.1  \\ \cline{2-16} 
\multicolumn{1}{|l|}{}                            & \multicolumn{1}{l|}{Supervised RegNet} & \multicolumn{1}{l|}{Supervised} & 43.74 & \multicolumn{1}{l|}{Supervised RegNet}   & \multicolumn{1}{l|}{Supervised} & 39.97 & \multicolumn{1}{l|}{Supervised RegNet}   & \multicolumn{1}{l|}{Supervised} & 38.12 & \multicolumn{1}{l|}{Supervised RegNet}   & \multicolumn{1}{l|}{Supervised} & 34.34 & \multicolumn{1}{l|}{Supervised RegNet}   & \multicolumn{1}{l|}{Supervised} & 29.28 \\ \hline \hline
\multicolumn{1}{|l|}{\multirow{3}{*}{Food}}       & \multicolumn{1}{l|}{SWaV (RN50w2)}     & \multicolumn{1}{l|}{SSL (ID)}   & 75.86 & \multicolumn{1}{l|}{SWaV (RN50w2)}       & \multicolumn{1}{l|}{SSL (ID)}   & 71.75 & \multicolumn{1}{l|}{DINO (ViT)}          & \multicolumn{1}{l|}{SSL (ID)}   & 68.06 & \multicolumn{1}{l|}{DINO (ViT)}          & \multicolumn{1}{l|}{SSL (ID)}   & 62.65 & \multicolumn{1}{l|}{DINO (ViT)}          & \multicolumn{1}{l|}{SSL (ID)}   & 54.52 \\ \cline{2-16} 
\multicolumn{1}{|l|}{}                            & \multicolumn{1}{l|}{DINO (ViT)}        & \multicolumn{1}{l|}{SSL (ID)}   & 74.47 & \multicolumn{1}{l|}{DINO (ViT)}          & \multicolumn{1}{l|}{SSL (ID)}   & 71.33 & \multicolumn{1}{l|}{SWaV (RN50w2)}       & \multicolumn{1}{l|}{SSL (ID)}   & 67.01 & \multicolumn{1}{l|}{SWaV (RN50w2)}       & \multicolumn{1}{l|}{SSL (ID)}   & 60.2  & \multicolumn{1}{l|}{SWaV (RN50w2)}       & \multicolumn{1}{l|}{SSL (ID)}   & 51.51 \\ \cline{2-16} 
\multicolumn{1}{|l|}{}                            & \multicolumn{1}{l|}{SWaV (RN50)}       & \multicolumn{1}{l|}{SSL (ID)}   & 72.27 & \multicolumn{1}{l|}{DINO (RN50)}         & \multicolumn{1}{l|}{SSL (ID)}   & 69.24 & \multicolumn{1}{l|}{DINO (RN50)}         & \multicolumn{1}{l|}{SSL (ID)}   & 65.59 & \multicolumn{1}{l|}{DINO (RN50)}         & \multicolumn{1}{l|}{SSL (ID)}   & 59.53 & \multicolumn{1}{l|}{DINO (RN50)}         & \multicolumn{1}{l|}{SSL (ID)}   & 50.93 \\ \hline
\end{tabular}
}
\end{table*}

Table \ref{tbl:homog_results} shows the linear evaluation performance for varying homography strengths for the benchmark datasets (see Appendix A.2.1 for more results). The supervised baselines perform best on CIFAR10 and CIFAR100 - two of the most common benchmarks in computer vision. However, the SSL ID models dominate the remaining 5 benchmark datasets. These 5 benchmarks contain more variety, and arguably serve as a more comprehensive evaluation than the CIFAR datasets, due to the much higher resolution inputs, higher cardinality label sets, and higher frequency differences between classes. For example, in the most challenging benchmark, Aircraft, there is not a single supervised model in the top 3 for any of the homography strengths.

We note that backbones based on the ViT \citep{vit} architecture (both SS and supervised) perform well, and are the best performing models for the majority of the datasets. This motivates further research into ViT-based SS models. Further, we note that wider ResNets (ResNet50w2) tend to perform better than narrower ones (ResNet50), as evidenced in the Color, Caltech101, and Food datasets. This is a known result in SSL for the usual linear evaluation strategy for these benchmarks, and our findings here suggest that this result generalises to the synthetic multi-view setting. It is also interesting to note that no pretext-task-based SSL model (SSL PT) appears in the top 3 for any of the datasets. This further motivates the recent focus of the research community on SSL ID methods, as they typically outperform SSL PT methods on all evaluated tasks. Interestingly, this general trend of SSL (ID) methods performing best, followed by Supervised (with SSL (PT) performing worst) holds for the \emph{bounded} homography as well (refer to Appendix A.2.2 for the full table of results).

\begin{table}[!h]
\caption{\label{tbl:reldec_homog_res}Relative decrease in performance from $H_{0.0}$ to $H_{0.2,0.4,0.6,0.8}$ for both the default and bounded homographies by model type. Results are averaged over the 7 benchmark datasets.}
\centering
\begin{tabular}{|c|l|c||c|}
\hline
\multicolumn{1}{|l|}{Decrease Strength} & Type       & Default Homography & Bounded Homography \\ \hline
\multirow{3}{*}{$H_{0.0} \rightarrow H_{0.2}$}                      & SSL (ID)   & -0.105             & \textbf{-0.0337}   \\ \cline{2-4} 
                                        & SSL (PT)   & -0.2547            & -0.1483            \\ \cline{2-4} 
                                        & Supervised & \textbf{-0.0725}   & -0.0704            \\ \hline \hline
\multirow{3}{*}{$H_{0.0} \rightarrow H_{0.4}$}                      & SSL (ID)   & -0.1797            & \textbf{-0.0743}   \\ \cline{2-4} 
                                        & SSL (PT)   & -0.3363            & -0.2646            \\ \cline{2-4} 
                                        & Supervised & \textbf{-0.1226}   & -0.1168            \\ \hline \hline
\multirow{3}{*}{$H_{0.0} \rightarrow H_{0.6}$}                      & SSL (ID)   & -0.2644            & \textbf{-0.1974}   \\ \cline{2-4} 
                                        & SSL (PT)   & -0.3806            & -0.4109            \\ \cline{2-4} 
                                        & Supervised & \textbf{-0.1932}   & -0.2252            \\ \hline \hline
\multirow{3}{*}{$H_{0.0} \rightarrow H_{0.8}$}                      & SSL (ID)   & -0.3485            & \textbf{-0.3933}   \\ \cline{2-4} 
                                        & SSL (PT)   & -0.4134            & -0.5642            \\ \cline{2-4} 
                                        & Supervised & \textbf{-0.2829}   & -0.4118            \\ \hline
\end{tabular}
\end{table}

Table \ref{tbl:reldec_homog_res} shows the \textbf{relative decrease} in performance by model type for both the default and bounded homographies. Interestingly, it seems the black background affects the results significantly because Supervised models perform best on all decrease strengths with the default homography (i.e. with the black background), but SSL (ID) models consistently perform best for all decrease strengths with the bounded homography (where no black background is present to bias results). This suggests that the supervised models are somewhat reliant on the black background of the homography to retain the performance as the homography gets stronger. However, without the black background, the supervised models struggle to retain their performance with stronger homographies. In contrast, the SSL (ID) models retain accuracy better when no black background is present to potentially bias the model. Unsurprisingly, SSL (PT) perform worst in both cases.

These results suggest that the SS techniques are more robust, in general, to changes in the perspective/viewpoint of the object \emph{as approximated by a homography}. This motivates further analysis of these methods on real-world, multi-view datasets to test in a more realistic scenario where viewpoint variation of the objects is more natural than a homography. We conduct these experiments and present the results in the next section.

\subsection{Multi-View Performance in-the-Wild} \label{sec:real_world}
Our motivation for this section is to benchmark the performance of SS representations against supervised alternatives for multi-view performance \emph{in-the-wild}. Previous work has not focused on the multi-view efficacy of SSL on such varied and real-world datasets, nor in this detail. We do this by running nearest neighbour searches (NNS) directly in the respective network's embedding spaces. We posit this is a very effective method of gauging the network embedding's multi-view invariance and robustness to viewpoint variation in the wild (i.e without adapting the weights or fine-tuning to a particular dataset). This set of experiments essentially serves as a realistic setting for analysing and measuring real-world viewpoint invariance.

\begin{table}[!h]
\centering
\caption{\label{tbl:top_k_real_world}Top 3 performing models for each real-world, multi-view dataset. Results are ranked.}
\scalebox{0.8}{
\begin{tabular}{|l|l|l|l|}
\hline
Dataset                      & Model             & Type       & Accuracy \\ \hline
\multirow{3}{*}{ALOI}        & SWaV (RN50w2)     & SSL (ID)   & 96.8     \\ \cline{2-4} 
                             & SWaV (RN50)       & SSL (ID)   & 96.22    \\ \cline{2-4} 
                             & DINO (RN50)       & SSL (ID)   & 96.16    \\ \hline\hline
\multirow{3}{*}{MVMC}        & RotNet (RN50)     & SSL (PT)   & 93.14    \\ \cline{2-4} 
                             & MoCov2 (RN50)     & SSL (ID)   & 92.3     \\ \cline{2-4} 
                             & Supervised (ViT)  & Supervised & 90.89    \\ \hline\hline
\multirow{3}{*}{MVC}     & SWaV (RN50w2)     & SSL (ID)   & 97.2     \\ \cline{2-4} 
                             & Jigsaw (RN50)     & SSL (PT)   & 95.11    \\ \cline{2-4} 
                             & Supervised (ViT)  & Supervised & 94.56    \\ \hline\hline
\multirow{3}{*}{Recon3D}     & Supervised (RN50) & Supervised & 94.94    \\ \cline{2-4} 
                             & Supervised RegNet & Supervised & 93.2     \\ \cline{2-4} 
                             & DINO (ViT)        & SSL (ID)   & 93.2     \\ \hline\hline
\multirow{3}{*}{Stereo Face} & SWaV (RN50w2)     & SSL (ID)   & 97.72    \\ \cline{2-4} 
                             & DINO (ViT)        & SSL (ID)   & 97.71    \\ \cline{2-4} 
                             & DINO (RN50)       & SSL (ID)   & 97.2     \\ \hline
\end{tabular}
}
\end{table}

Table \ref{tbl:top_k_real_world} shows the top 3 models for each of the real-world datasets (see Appendix B.1 for more results). Clearly, SS representations dominate in terms of overall performance on these datasets. Interestingly, the ViT models are not as effective in this setting as compared to the synthetic setting with the homographies. Further, the pretext-task-based SSL methods appear more frequently in this setting, which suggests that the representations learned with this paradigm contain information useful for encoding object viewpoint (more so than can be leveraged in the synthetic setup in the previous section).

It should be noted, however, that the supervised models perform best on the Recon3D dataset. This is somewhat expected since this dataset contains images with a significant distribution shift from ImageNet. It has been shown in previous work that SSL models are less robust than supervised models (pre-trained on ImageNet) when evaluated on datasets with a large distribution shift from ImageNet \citep{Ericsson2021HowTransfer} (e.g. non-natural images), and this experiment replicates these findings in the multi-view setting.

\begin{table}[!h]
\centering
\caption{\label{tbl:real_world_model_type}Real-world multi-view dataset performance by model type.}
\begin{tabular}{l|c|c|c|c|c|}
\cline{2-6}
                                 & \multicolumn{1}{l|}{ALOI} & \multicolumn{1}{l|}{MVC} & \multicolumn{1}{l|}{MVMC} & \multicolumn{1}{l|}{Recon3D} & \multicolumn{1}{l|}{Stereo Face} \\ \hline
\multicolumn{1}{|l|}{SSL (ID)}   & \textbf{95.66}            & \textbf{92.73}           & \textbf{90.41}            & 88.34                        & \textbf{94.36}                   \\ \hline
\multicolumn{1}{|l|}{SSL (PT)}   & 84.04                     & 88.47                    & 85.21                     & 63.26                        & 70.83                            \\ \hline
\multicolumn{1}{|l|}{Supervised} & 93.58                     & 89.32                    & 88.34                     & \textbf{90.72}               & 91.56                            \\ \hline
\end{tabular}
\end{table}

Table \ref{tbl:real_world_model_type} provides a summary of performance on these datasets by model type. These results are computed by averaging over all trials for each dataset. SSL (ID) models perform best overall except on Recon3D. Supervised performs 2nd best on average for all datasets, and SSL (PT) models are 3rd by some distance on average. We delve further into these results below, going into more granular detail for each dataset. Please note that for all plots below, SSL (ID), SSL (PT), and Supervised models are visualised by red, green, and grey tones, respectively.

\begin{figure}[!h]
    \centering
    \includegraphics[width=\columnwidth]{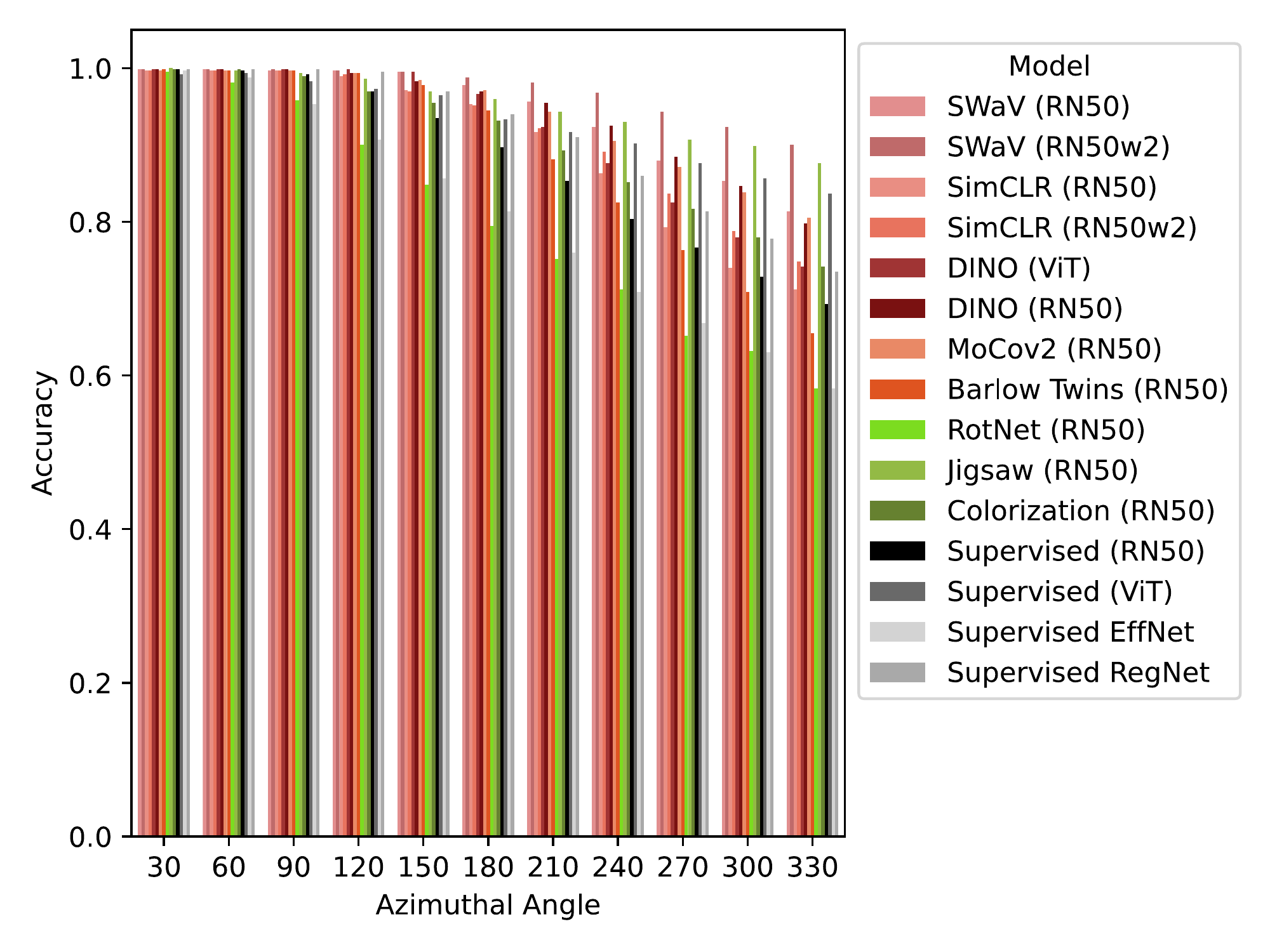}
    \caption{NNS results on the MVC dataset. The $x$-axis represents the difference in azimuthal angle between the training view and testing view.}
    \label{fig:knn_results}
\end{figure}

Figure \ref{fig:knn_results} shows the performance of such an NNS on a multi-view car dataset for the various models at different azimuthal angle differences. To obtain these metrics, we average the testing accuracy over all possible pairs of images that are $30 n$ degrees apart ($n \in \{1, 2, \dots, 11\}$). In this case, for each trial, the training and testing dataset both have a cardinality equal to the number of classes (i.e. in this case, the number of unique cars in the dataset). The reported results are an average of 30 randomised trials of different images at these angle differences.

On average, SSL (ID) models perform best, with the SWaV models performing particularly well. The Jigsaw SSL (PT) model performs surprisingly well on this task. It is clear that as the azimuthal angle increases, performance decreases for all models. Moreover, at the large azimuthal angle differences (i.e. $> 150$, when the task is more difficult), the performance improvement of SSL models vs the supervised baselines is notably larger. The Supervised (ViT) model consistently performs best out of all the supervised baselines by a large margin, and indeed performs similarly to some of the better SSL models.

\begin{figure}[!h]
    \centering
    \includegraphics[width=\columnwidth]{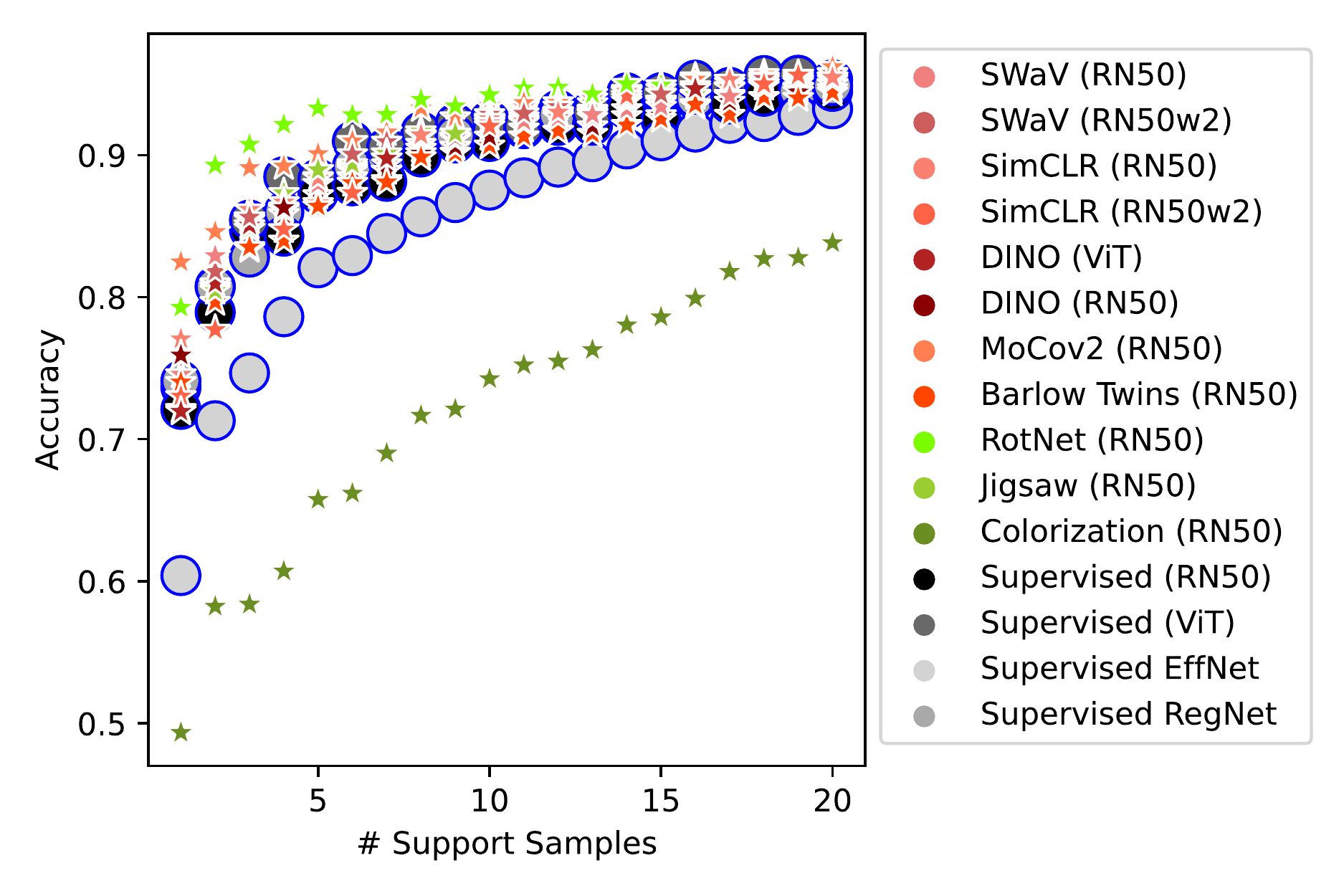}
    \caption{NNS results MVMC dataset. The stars denote SSL models (both ID and PT), and circles with blue borders denote supervised models.}
    \label{fig:mvmc_results}
\end{figure}

Figure \ref{fig:mvmc_results} shows the results of an NNS on a real-world, multi-view, multi-class dataset \citep{mvmc}. We vary the number of support samples - that is, the number of images of each class in the training set. Each point represents a mean of 100 random trials. We see that with as few as \emph{one} image per class, the SS techniques outperform the supervised baselines significantly. We note that with a lower number of support samples (e.g. 1-2 samples), the variance of the performance metric is fairly high, which motivates the large number of trials. This is due to the fact that the performance depends on whether that single support sample per class is a canonical example of that class, or a poor representative sample that makes classification difficult. Evaluating with a small number of support samples enables us to truly test the efficacy of the embeddings for multi-view object recognition.

As the number of support samples increases, the performance gap between SSL and supervised naturally decreases. This shows that when enough context is available, the supervised baseline improves - most likely due to the support samples illustrating a large enough diversity of different views, and at that stage the viewpoint robustness becomes unnecessary. However, the superior performance of SSL when the amount of context is very low suggests that SSL embeddings encode more information relevant to multi-view tasks. We posit that this is due to both the better occlusion invariance of SSL techniques and the nature of the ID objective in general.

\begin{figure}[!h]
    \centering
    \includegraphics[width=\columnwidth]{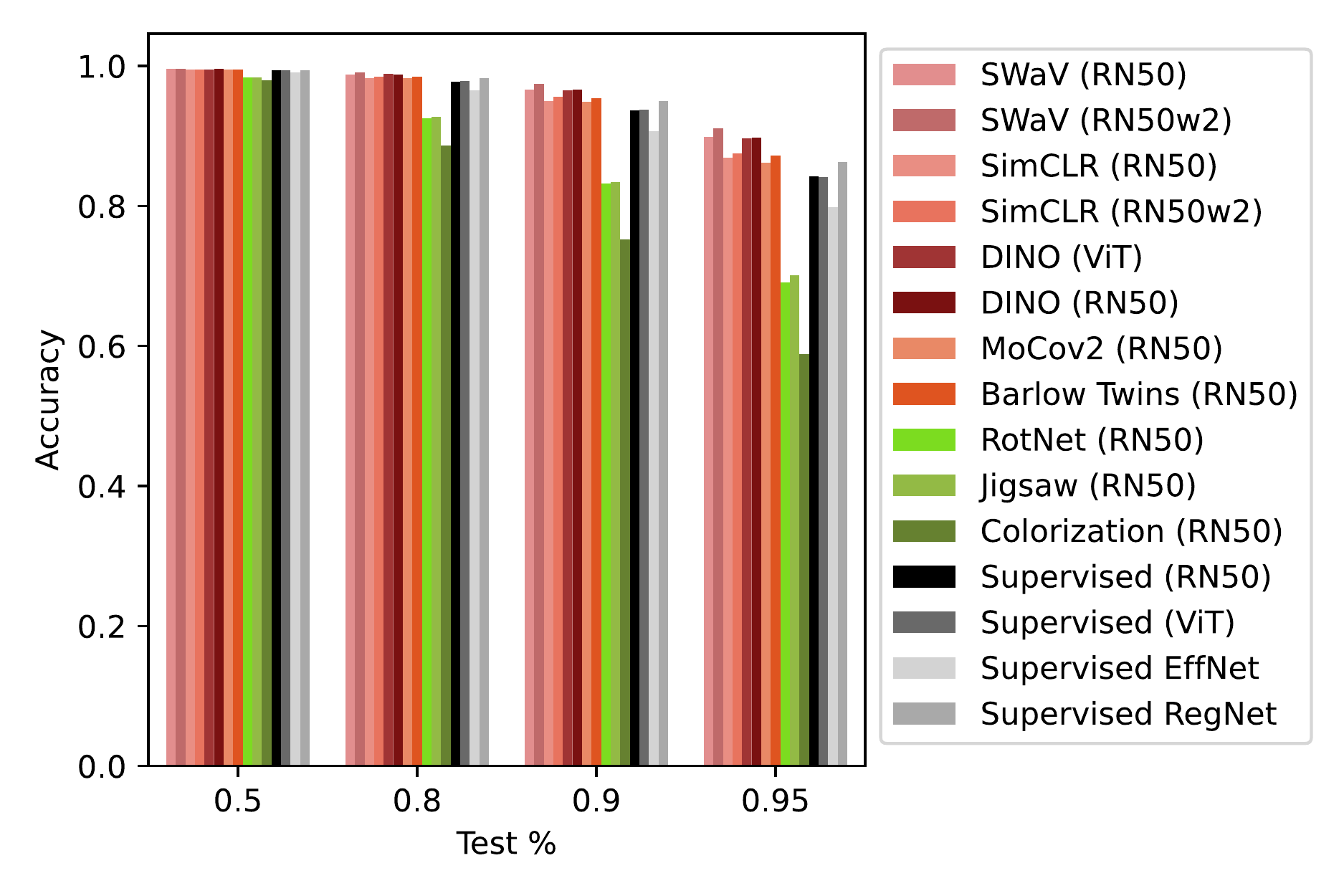}
    \caption{NNS results on the ALOI dataset.}
    \label{fig:aloi_results}
\end{figure}

Finally, we report results on the ALOI dataset (Fig. \ref{fig:aloi_results}). We evaluate by increasing the test split percentage (up to 95\% of the data for testing), essentially giving the models less and less context when performing the nearest neighbour search to test how effective the representations are at detecting the objects at different viewpoints. Similarly to the MVMC dataset above, when the amount of context provided to the model is reduced, the performance difference between SS and supervised models increases on average. Furthermore, SSL (PT) methods are consistently outperformed by the two other model types. This, as well as the results of SSL (PT) on the other real-world, multi-view dataset, suggests that the pretext tasks in the literature are not conducive to learning representations that are useful for reliable multi-view recognition, and do not encode viewpoint as reliable as the SSL ID paradigm.

\subsection{Self-Supervised Representations vs. Supervised Representations}
It is clear from Sections \ref{sec:synthetic} and \ref{sec:real_world} that representations learned through self-supervision perform better, on average, than representations learned with supervision for the task of multi-view detection (when the models are pre-trained on ImageNet). The results suggests that this claim holds for both controlled distortion to object viewpoint through homographies (Section \ref{sec:synthetic}), as well as in real-world experiments on datasets that are inherently multi-view (Section \ref{sec:real_world}). In particular, SSL (ID) models are particularly effective, which mirrors similar findings outside the multi-view domain in the SSL literature \citep{Ericsson2021HowTransfer}.

We posit that ID models are so effective at multi-view object recognition for one main reason. A typical ID loss is of the following form: $L(\mathbf{x}, \mathbf{y})$, where $\mathbf{x}$ and $\mathbf{y}$ are the embeddings from the network for two distinct \emph{views} of the \textbf{same} input image. These views are typically generated through a series of data augmentations, such as random cropping, colour distortion, and random flipping. As the network is trained, these two vectors are learned to be represented by the same vector. Since the two views represent different parts (i.e. crops) of the image, these may be different parts of the same object (particularly in object-centric datasets that are the norm in the SSL literature). This essentially means that the network is explicitly trained to be able to recognise an object from potentially very little context such as a random crop. This may explain the superior performance of SSL in general, particularly for datasets such as MVC and MVMC, as well as why these SSL (ID) models work especially well with very little context when compared to the various supervised baselines. Similar intuitions have been found in previous work in measuring invariances of SSL models, where they were shown to be more occlusion invariant than supervised alternatives \citep{demyst}. It should be noted, that the performance benefits of SS learning decreases as the amount of context provided to the networks increases (see Figures \ref{fig:aloi_results}, \ref{fig:knn_results}, and \ref{fig:mvmc_results}). This is expected, as the networks have sufficient information about the objects when a large number of embeddings are available to perform the NNS or classification. It is, however, particularly useful to have good performance in a low-data regime, when little context is available. This is the main advantage of SS representations for multi-view object recognition.

Interestingly, some known results from outside the multi-view domain still hold. Performance of SSL models on data that has a large distribution shift from ImageNet has been shown to be lower than supervised learning \citep{Ericsson2021HowTransfer}, and those findings generalise to our multi-view setting, as evidenced by our results on the Recon3D dataset. This is likely due to the benefits of the ID objective discussed above not being able to generalise beyond ImageNet-like images, and not being able outweigh the benefits of a strong supervision signal from human-generated labels in this setting.

Lastly, ViT based models have shown promising results in SSL literature \citep{dino}, and it is clear from our results that ViTs perform particularly well at multi-view recognition (both SS and supervised versions). We suggest that this is due to the fact (in the SS case) that the SS transformers automatically discover objects (as shown by \citet{dino} by visualising the attention maps of the ViT-based DINO, where the model learns an approximate semantic segmentation of the object). We posit that this property of ViTs assists in multi-view object recognition, and encourage the representations learned by these models to be more viewpoint invariant.

\section{Conclusion}
\label{sec:conclusion}
We observe that in multiple different scenarios, representations learned through modern self-supervision are more robust than supervised learning with respect to viewpoint changes. We find this holds both for approximate viewpoint variation with homographies, as well as real-world viewpoint variation on the numerous inherently multi-view datasets. Interestingly, even with very little context (see Figures \ref{fig:knn_results}, \ref{fig:mvmc_results}, and \ref{fig:aloi_results}), or in some cases with large context (see Table \ref{tbl:homog_results}), SSL consistently outperforms supervised learning in the multi-view setting.

We posit that these findings suggest that SS representations encode information more pertinent to object parts, which enables improved robustness to viewpoint. This is likely a byproduct of the fact that most modern SSL models (such as SWaV, DINO, and SimCLR) are trained with strong data augmentation (e.g. random crops) and ID, which encourages this property of being able to recognise objects from a comparatively smaller context such as object parts.

The use cases for these models and associated results are diverse. Models based on SSL should be preferred to supervised alternatives when annotation is a bottleneck/prohibitive. Further, SSL models are preferable when the application has a need to be able to viewpoint invariant, or to recognise a set of distinct objects at different viewpoints - \emph{particularly in low-data regimes}. Many such applications exist, since viewpoint invariance is a core component of effective real-world computer vision \citep{demyst}. The supervised models should be preferred when there is no guarantee that the downstream application will consist of images similar to that of the pretraining data. Both previous work, and this work, have shown that supervised models outperform SS models in different contexts with these sorts of data distribution shifts.

\bibliographystyle{unsrtnat}
\bibliography{references}

\newpage
\appendix
\section{Robustness Deformation}
\label{apx:robust_defor}

\subsection{Data Augmentation Details}
For NNS and linear evaluation experiments, the training transformation involves resizing the image such that the smaller side has length $224$, and then performing a $224 \times 224$ centre crop (hereafter referred to as the RCC). The testing transform for these experiments is the RCC transformation, followed by an optional homography $H_{\alpha}$. For the fine-tuning experiments, the training transform is a random-resized crop, and the testing transform is the same as the NNS and linear evaluation paradigms.

\newpage
\subsection{Linear Evaluation}
\subsubsection{Default Homography}
\begin{table}[!h]
\caption{\label{tbl:homog_results_ci}Linear evaluation performance on common image classification benchmarks for increasing homography strengths. Results are averaged over 10 random trials, and models are ranked.}
\scalebox{0.49}{
\begin{tabular}{l|lll|lll|lll|lll|lll|}
\cline{2-16}
                                                   & \multicolumn{3}{c|}{$H_{0.0}$}                                                            & \multicolumn{3}{c|}{$H_{0.2}$}                                                            & \multicolumn{3}{c|}{$H_{0.4}$}                                                            & \multicolumn{3}{c|}{$H_{0.6}$}                                                            & \multicolumn{3}{c|}{$H_{0.8}$}                                                            \\ \hline
\multicolumn{1}{|l|}{\multirow{15}{*}{CIFAR10}}    & \multicolumn{1}{l|}{Supervised (ViT)}    & \multicolumn{1}{l|}{Supervised} & 92.29 & \multicolumn{1}{l|}{Supervised (ViT)}    & \multicolumn{1}{l|}{Supervised} & 91.68 & \multicolumn{1}{l|}{Supervised (ViT)}    & \multicolumn{1}{l|}{Supervised} & 90.94 & \multicolumn{1}{l|}{Supervised (ViT)}    & \multicolumn{1}{l|}{Supervised} & 88.7  & \multicolumn{1}{l|}{Supervised (ViT)}    & \multicolumn{1}{l|}{Supervised} & 79.94 \\ \cline{2-16} 
\multicolumn{1}{|l|}{}                             & \multicolumn{1}{l|}{SWaV (RN50w2)}       & \multicolumn{1}{l|}{SSL (ID)}   & 91.52 & \multicolumn{1}{l|}{DINO (ViT)}          & \multicolumn{1}{l|}{SSL (ID)}   & 89.04 & \multicolumn{1}{l|}{DINO (ViT)}          & \multicolumn{1}{l|}{SSL (ID)}   & 83.67 & \multicolumn{1}{l|}{Supervised EffNet}   & \multicolumn{1}{l|}{Supervised} & 74.87 & \multicolumn{1}{l|}{Supervised EffNet}   & \multicolumn{1}{l|}{Supervised} & 67.83 \\
\hline
\end{tabular}
}
\end{table}

\newpage
\subsubsection{Bounded Homography}
\begin{table}[!h]
\caption{\label{tbl:bounded_homog_results}Linear evaluation performance on common image classification benchmarks for increasing homography strengths. Models are ranked.}
\scalebox{0.5}{
\begin{tabular}{l|lll|lll|lll|lll|}
\cline{2-13}
                                                   & \multicolumn{3}{c|}{$H_{0.2}$}                                                         & \multicolumn{3}{c|}{$H_{0.4}$}                                                         & \multicolumn{3}{c|}{$H_{0.6}$}                                                         & \multicolumn{3}{c|}{$H_{0.8}$}                                                         \\ \hline
\multicolumn{1}{|l|}{\multirow{15}{*}{CIFAR10}}    & \multicolumn{1}{l|}{Supervised (ViT)}    & \multicolumn{1}{l|}{Supervised} & 91.01 & \multicolumn{1}{l|}{Supervised (ViT)}    & \multicolumn{1}{l|}{Supervised} & 88.76 & \multicolumn{1}{l|}{Supervised (ViT)}    & \multicolumn{1}{l|}{Supervised} & 81.44 & \multicolumn{1}{l|}{Supervised (ViT)}    & \multicolumn{1}{l|}{Supervised} & 65.78 \\ \cline{2-13} 
\multicolumn{1}{|l|}{}                             & \multicolumn{1}{l|}{SWaV (RN50w2)}       & \multicolumn{1}{l|}{SSL (ID)}   & 90.06 & \multicolumn{1}{l|}{DINO (ViT)}          & \multicolumn{1}{l|}{SSL (ID)}   & 87.17 & \multicolumn{1}{l|}{DINO (ViT)}          & \multicolumn{1}{l|}{SSL (ID)}   & 78.66 & \multicolumn{1}{l|}{DINO (ViT)}          & \multicolumn{1}{l|}{SSL (ID)}   & 61.9  \\ \cline{2-13} 
\multicolumn{1}{|l|}{}                             & \multicolumn{1}{l|}{DINO (ViT)}          & \multicolumn{1}{l|}{SSL (ID)}   & 89.62 & \multicolumn{1}{l|}{SWaV (RN50w2)}       & \multicolumn{1}{l|}{SSL (ID)}   & 85.67 & \multicolumn{1}{l|}{Barlow Twins (RN50)} & \multicolumn{1}{l|}{SSL (ID)}   & 72.97 & \multicolumn{1}{l|}{SWaV (RN50w2)}       & \multicolumn{1}{l|}{SSL (ID)}   & 54.81 \\
\hline
\end{tabular}
}
\end{table}

\newpage
\section{Real-World Datasets}
\subsection{Overall Performance}
\begin{table}[!h]
\centering
\caption{\label{tbl:knn_results_full}KNN performance on the real-world, multi-view benchmarks. Models are ranked.}
\scalebox{0.65}{
\begin{tabular}{l|l|l|l|}
\cline{2-4}
                                                    & Model               & Type       & Accuracy \\ \hline
\multicolumn{1}{|l|}{\multirow{15}{*}{ALOI}}        & SWaV (RN50w2)       & SSL (ID)   & 96.8     \\ \cline{2-4} 
\multicolumn{1}{|l|}{}                              & SWaV (RN50)         & SSL (ID)   & 96.22    \\ \cline{2-4} 
\multicolumn{1}{|l|}{}                              & DINO (RN50)         & SSL (ID)   & 96.16    \\ \cline{2-4} 
\multicolumn{1}{|l|}{}                              & DINO (ViT)          & SSL (ID)   & 96.12    \\ \cline{2-4} 
\multicolumn{1}{|l|}{}                              & SimCLR (RN50w2)     & SSL (ID)   & 95.27    \\ \cline{2-4} 
\multicolumn{1}{|l|}{}                              & Barlow Twins (RN50) & SSL (ID)   & 95.1     \\ \cline{2-4} 
\multicolumn{1}{|l|}{}                              & SimCLR (RN50)       & SSL (ID)   & 94.91    \\
\hline
\end{tabular}
}
\end{table}

\end{document}